# Validation Rules for Assessing and Improving SKOS Mapping Quality

Hong Sun, Jos De Roo, Marc Twagirumukiza, Giovanni Mels,

Kristof Depraetere, Boris De Vloed, Dirk Colaert

*Advanced Clinical Applications Research Group, Agfa HealthCare, Gent, Belgium*
`{hong.sun, jos.deroo, marc.twagirumukiza, giovanni.mels, kristof.depraetere, boris.devloed, dirk.colaert}@agfa.com`

**Abstract.** The Simple Knowledge Organization System (SKOS) is popular for expressing controlled vocabularies, such as taxonomies, classifications, etc., for their use in Semantic Web applications. Using SKOS, concepts can be linked to other concepts and organized into hierarchies inside a single terminology system. Meanwhile, expressing mappings between concepts in different terminology systems is also possible. This paper discusses potential quality issues in using SKOS to express these terminology mappings. Problematic patterns are defined and corresponding rules are developed to automatically detect situations where the mappings either result in 'SKOS Vocabulary Hijacking' to the source vocabularies or cause conflicts. An example of using the rules to validate sample mappings between two clinical terminologies is given. The validation rules, expressed in N3 format, are available as open source.

## 1 Introduction

The Simple Knowledge Organization System (SKOS) [1] [2] provides a data model and vocabulary for expressing controlled vocabularies such as taxonomies, classifications, etc. Many organizations have published their controlled vocabularies using SKOS for their use in Semantic Web applications [5]. Representing terminology systems with SKOS provides considerable benefits in formalizing and sharing data by means of representing the concepts, as well as their relations, in a common and machine readable way. In order to further facilitate the knowledge transfer between data publisher and data consumer, terminology mapping is urgently required. Such a requirement is particularly prominent in the clinical domain, where many clinical terminology systems are in use. Mappings between these different terminologies become a prerequisite for clinical data sharing between applications. For example, a hospital uses the ICD-10[1] coding system to record diagnosis; however, for reporting adverse drug events (ADE) detected in this hospital, it is mandatory to use MedDRA[2] codes to describe the adverse drug events. Mappings between ICD-10 terms and MedDRA terms are therefore indispensable in the discussed example for ADE reporting.

Terminology mappings between different terminology systems do exist in many domains. For example, in the clinical domain, the International Health Terminology Standards Development Organisation, together with the World Health Organization, developed the SNOMED CT[3] to ICD-10 mapping [3], where the mappings are expressed in an Excel style sheet. In the OMOP project [4], SNOMED CT to ICD-9-CM mappings are stored in a relational table. However, most of the mappings are expressed in a non-semantic format, which prevents their direct use in semantic web applications. Representing terminology mappings in a semantic format is therefore required so as to utilize these mappings in semantic web applications.

The SKOS specification defines five mapping properties: skos:broadMatch, skos:narrowMatch, skos:relatedMatch, skos:closeMatch, and skos:exactMatch, to express mappings between concepts in different schemes. These mapping properties allow specifying mappings in different situations (e.g. broader, narrow, etc.).

However, we discovered that transitive features can be inferred from the SKOS mapping properties, which may assert unintended semantic relations to source vocabularies. Such injected relations are not developed by the owners of source vocabularies and therefore we consider such assertion as 'SKOS Vocabulary Hijacking' [5]. Furthermore, as such assertions mostly exist implicitly and only become explicit after inference, the mapping creators might not be aware of such side effects brought by their mappings, which may lead to serious consequences. Besides the consequence of resulting in 'SKOS Vocabulary Hijacking', the mappings may also cause conflicts after applying the inference allowed by the SKOS specification.

This paper first analyzes the cause of the above mentioned issues, and then explicitly states seven patterns that may cause those issues. Validation rules to detect each of those patterns are also presented in N3[4] format. The resources for a test case on sample mappings are also provided.

## 2   SKOS Vocabulary Hijacking Caused by SKOS Mapping

In [5], the concept 'SKOS vocabulary hijacking – the assertion of facts about vocabularies published by others' is coined, mimicking the term 'ontology hijacking - the redefinition by third parties of external classes/properties such that reasoning over data using those external terms is affected' [6]. The SKOS authors permit such assertion and considered such scenario as 'asserting semantic links within someone else's concept scheme (a.k.a. "kos enrichment")'[5], and suggest to 'use graph provenance to distinguish between "authoritative" and "third party" assertions'.

---

[1] International statistical classification of disease and related health problems, Tenth Revision (ICD-10)
[2] The Medical Dictionary for Regulatory Activities (MedDRA): http://www.meddra.org/
[3] SNOMED Clinical Terms: http://www.ihtsdo.org/snomed-ct/
[4] Notation 3: http://www.w3.org/TeamSubmission/n3/

Although the above mentioned intentional assertion is conditionally allowed, we deem the unintentional assertions brought by SKOS mapping as a different scenario which may bring serious consequence. The mapping creators should be notified when these mappings are resulting in these inferred relation assertions.

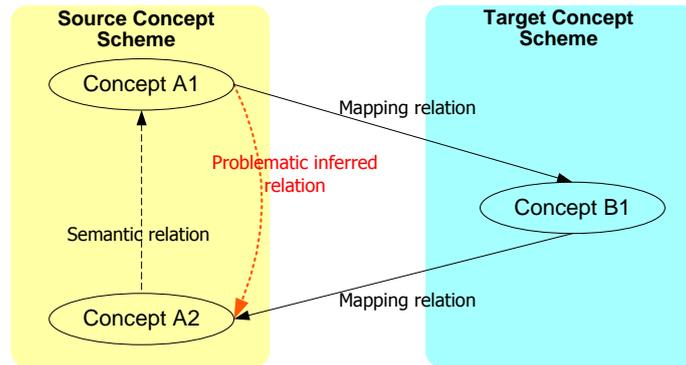

Figure 1. Basic vocabulary hijacking pattern

Figure 1 shows the basic pattern of SKOS vocabulary hijacking resulting from SKOS mappings. As most of the SKOS mapping properties (or relations inferred from mapping relations) are symmetric and transitive (see Listing 1 and Listing 2), it is possible to infer a relation (as displayed by the red dotted line in Figure 1) between two concepts inside a same concept scheme. If such an inferred relation is not stated in (or cannot be inferred from) that concept scheme, then inferring such a relation via SKOS mapping relations is considered as vocabulary hijacking.

In addition, if the inferred relation (from the mapping relations) is contradictory to an existing relation, e.g. the semantic relations displayed as black dashed line in Figure 1, it would be considered as a conflict, and the related mappings are considered to be erroneous and need to be corrected.

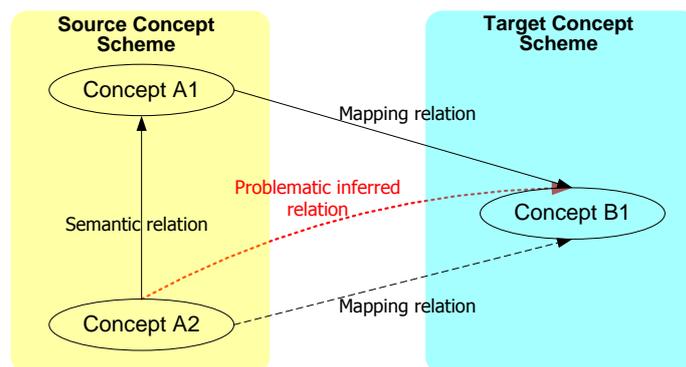

Figure 2. Conflicts with mapping relations

---
[5] http://www.w3.org/2006/07/SWD/wiki/WashingtonAgenda/MappingIssues

Furthermore, problematic relations can also be inferred from the combination of semantic relations and mapping relations. Figure 2 shows such a pattern where a problematic relation is inferred (as displayed by the red dotted line) from the semantic relation between A2 and A1 together with the mapping relation between A1 and B1. If this inferred relation is contradictory to an existing relation, e.g. the mapping relations between A2 and B1 (what displayed as black dashed line in Figure 2), it would be considered as a conflict.

The next section presents patterns that either result in vocabulary hijacking or creating conflicts. Rules that detect such patterns are also provided in N3 format.

## 3 Problematic Patterns in SKOS Mapping

This section describes the patterns that either result in vocabulary hijacking or create conflicts. Section 3.1 introduces a set of prerequisite rules to simplify the relations between concepts, as well as necessary inference to detect problematic patterns. Detailed descriptions of seven problematic patterns are given in Sections 3.2 to 3.8. N3 rules that detect the discussed patterns are published [7]. The structures of the problematic patterns are based on the assumption that the SKOS mapping and semantic relation properties are used following the conventional way: the relation between two concepts in different concept schemes is expressed with SKOS mapping properties, while the relation between two concepts within a same concept scheme is expressed with SKOS semantic relation properties.

### 3.1 Inference of Problematic Pattern

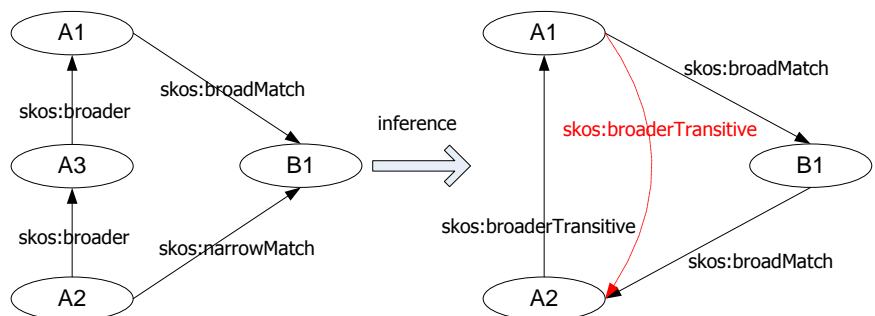

Figure 3. An example of inferring a problematic pattern

Figure 3 shows an example of inferring a problematic pattern in SKOS mapping. The left part of the figure shows the original SKOS concepts and relations where upon the SKOS mapping validation rules are to be applied. The concepts A1, A2, and A3 are in concept scheme A, their relations are expressed with skos:broader. B1 is a concept that belongs to concept scheme B, the relations between the concepts in scheme A and scheme B are expressed with skos:broadMatch and skos:narrowMatch. The listed mappings appear to be unidirectional and there is no obvious conflict.

A minimal set of SKOS inference rules as displayed in Listing 1 are applied to those original mappings as prerequisite in order to simplify the semantic and mapping relations. The inference rules are developed strictly following the SKOS specification. They are expressed in N3 format and executed by Euler YAP Engine (EYE) [8], an open source reasoning engine. After applying the rules in Listing 1, the retained semantic relations inside a concept scheme are skos:broaderTransitive and skos:related, where the skos:related is stated in both directions. The retained mapping relations among different concept schemes are skos:broadMatch, skos:exactMatch and skos:relatedMatch, where the latter two relations are stated in both directions.

Listing 1. Prerequisite Inference Rules

```
@prefix skos: <http://www.w3.org/2004/02/skos/core#>.

1  #inference on mapping relations
2  { ?x skos:exactMatch ?y } => { ?y skos:exactMatch ?x }.
3  { ?x skos:narrowMatch ?y } => { ?y skos:broadMatch ?x }.
4  { ?x skos:relatedMatch ?y } => { ?y skos:relatedMatch ?x }.
5
6  #inference on semantic relations
7  { ?x skos:narrower ?y } => { ?y skos:broader ?x }.
8  { ?x skos:related ?y } => { ?y skos:related ?x }.
9  { ?x skos:broader ?y } => { ?x skos:broaderTransitive ?y }.
10 { ?x skos:broaderTransitive ?y. ?y skos:broaderTransitive ?z } => { ?x skos:broaderTransitive ?z }.
```

After applying the inference stated in Listing 1, the pattern displayed on the left side of Figure 3 is translated into the pattern displayed on the right side, except for the red dotted relation (*A1 skos:broaderTransitive A2*). As can be observed, the relations represented in the right side pattern are reduced to skos:broadmatch and skos:broaderTransitive.

It can be observed that after applying the prerequisite inference rules, the unidirectional mapping relations as what represented on the left side of Figure 3 become bidirectional as what represented on the right side. Apply the rules in Listing 2 would further entail the relation (*A1 skos:broaderTransitive A2*) from the skos:broadMatch relations.

Listing 2. Deducing a Potential Vocabulary Hijacking Relation

```
{ ?x skos:broadMatch ?y } => { ?y skos:broader ?x }.
{ ?x skos:broader ?y } => { ?x skos:broaderTransitive ?y }.
{ ?x skos:broaderTransitive ?y. ?y skos:broaderTransitive ?z } => { ?x skos:broaderTransitive ?z }.
```

If the inferred relation is not stated in (or cannot be inferred from) concept scheme A, then inferring such a relation via SKOS mapping relations is considered as vocabulary hijacking of concept scheme A. What we consider as dangerous is that although such an inferred relation is brought by the mappings between concept scheme A and concept scheme B, the creator of the mappings might not be aware of such consequence. It is therefore important to pass the problematic patterns to the mapping creators for validation. In the remainder of Section 3, we present rules that detect and capture the problematic patterns.

### 3.2 PATTERN 1: Assert skos:broaderTransitive relation via skos:broadMatch mapping

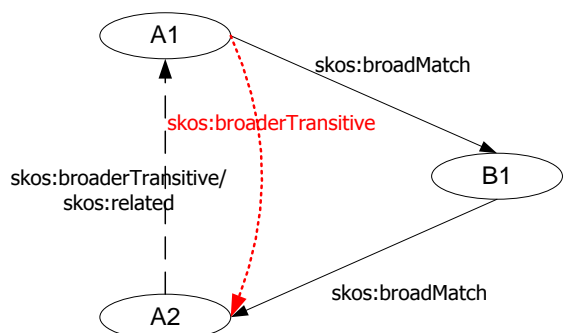

Figure 4. Validation pattern 1

Figure 4 shows the problematic pattern where a skos:broaderTransitive relation is asserted via the skos:broadMatch mappings. This problematic relation is displayed as red dotted lines. If such an inferred relation is not stated in (or cannot be inferred from) concept scheme A, then it is considered as a vocabulary hijacking. In addition, such an inferred relation may result in non-consistent if one of the relations represented in the black dashed line exists.

The rule displayed in Listing 3 detects the problematic vocabulary hijacking pattern and classifies this pattern in the output of the rule, so that the detected problematic patterns can be checked.

Listing 3. Pattern1VocabularyHijacking

```
@prefix skos: <http://www.w3.org/2004/02/skos/core#>.
@prefix validation: <http://eulersharp.sourceforge.net/2003/03swap/skos-mapping-validation-rules#>.
@prefix e: <http://eulersharp.sourceforge.net/2003/03swap/log-rules#>.
1 {
2      ?A1 skos:broadMatch ?B1.
3      ?B1 skos:broadMatch ?A2.
4      ?SCOPE e:findall ( ?A1 { ?A1 skos:broaderTransitive ?A2. } () ).
5      {      ?A1 skos:broadMatch ?B1.
6             ?B1 skos:broadMatch ?A2. } e:graphCopy ?pattern.
7 } => {
8      ?pattern a validation:Pattern1VocabularyHijacking.
9 }.
```

In Listing 3, Line 2-4 detect a problematic pattern which may assert a skos:broaderTransitive relation that is not existing in the source concept scheme before. Line 4 states the fact that the triple '?A1 skos:broaderTransitive ?A2' should not exist in the ?SCOPE. The ?SCOPE is bound to the deductive closure of all the input graphs. This is scoped negation as failure. Line 5 and 6 pass the detected problematic mappings to the graph ?pattern. The property e:graphCopy is used to have unification, so that repetitions are removed. The symbol '=>' stands for log:implies [9], its subject (the left side graph of '=>') is the antecedent graph, and the object (the right side graph) is the consequent graph. Line 8 is the result of this rule, where the detected pattern is considered an instance of the validation:Pattern1VocabularyHijacking class. A set of classes are defined corresponding to each problematic pattern. Both the problematic pattern classes and the detection rules are published [7].

Listing 4. Pattern1NonConsistentWithSKOSRules

```
1  {
2      ?A1 skos:broadMatch ?B1.
3      ?B1 skos:broadMatch ?A2.
4      ?A1 skos:related ?A2.
5      {   ?A1 skos:broadMatch ?B1.
6          ?B1 skos:broadMatch ?A2.
7          ?A1 skos:related ?A2. } e:graphCopy ?pattern.
8  } => {
9      ?pattern a validation:Pattern1NonConsistentWithSKOSRules.
10 }.
```

Other than vocabulary hijacking, if the inferred relations from the SKOS mapping relations are conflicting with the existing/inferred semantic relations inside the source or target concept scheme, the mappings would be considered as non-consistent. A set of N3 rules (SKOS-Rules[6]) has been developed where a set of false patterns are described strictly according to the SKOS specification. Listing 4 shows one of such non-consistent patterns, where the inferred (A1 skos:broaderTransitive A2) relation is considered as non-consistent with a (*A1 skos:related A2*) relation (according to SKOS Integrity Conditions S27).

Listing 5. Pattern1NonConsistentWithSKOSExtraRules

```
1  {
2      ?A1 skos:broadMatch ?B1.
3      ?B1 skos:broadMatch ?A2.
4      ?A2 skos:broaderTransitive ?A1.
5      {   ?A1 skos:broadMatch ?B1.
6          ?B1 skos:broadMatch ?A2.
7          ?A2 skos:broaderTransitive ?A1. } e:graphCopy ?pattern.
8  } => {
9      ?pattern a validation:Pattern1NonConsistentWithSKOSExtraRules.
10 }.
```

In addition, because of the SKOS specification makes the least ontological commitment, there are patterns although not listed as non-consistent in SKOS specification, but nevertheless are considered as conflicts by convention. The SKOS specification suggests end users to avoid such patterns, and we therefore consider such patterns problematic as well. A set of SKOS validation rules (SKOS-Extra-Rules[7]) are developed to describe the patterns that the SKOS specification considers as bad practice. Listing 5 describes one of such cases: following the transitive nature of skos:broaderTransitive, the inferred relation (*A1 skos:broaderTransitive A2*) together with the relation (*A2 skos:broaderTransitive A1*) stated in concept scheme A can deduce a new relation (*A2 skos:broaderTransitive A2*). Such a cyclic hierarchical relation is considered as a false pattern in the SKOS-Extra-Rules. Nevertheless, it is still up to the terminology system to define if such relations are illegal or not. The pattern listed in Listing 5 is anyhow considered as non-consistent with SKOS-Extra-Rules and is presented to the mapping creator for further validation.

---

[6] http://eulersharp.sourceforge.net/2003/03swap/skos-rules
[7] http://eulersharp.sourceforge.net/2003/03swap/skos-extra-rules

Listing 6. Classification of Problematic Patterns

```
@prefix validation: <http://eulersharp.sourceforge.net/2003/03swap/skos-mapping-validation-rules#>.
@prefix rdfs: <http://www.w3.org/2000/01/rdf-schema#>.

validation:Pattern1VocabularyHijacking   a rdfs:Class;
       rdfs:comment "The inferred skos:broaderTransitive relation via skos:broadMatch mapping is
           considered as vocabulary hijacking to the original vocabulary".
validation:Pattern1NonConsistentWithSKOSRules a rdfs:Class;
       rdfs:comment "The inferred skos:broaderTransitive relation via skos:broadMatch mapping is
           considered as contradictory with the existing skos:related relation".
validation:Pattern1NonConsistentWithSKOSExtraRules    a rdfs:Class;
       rdfs:comment "The inferred skos:broaderTransitive relation via skos:broadMatch  mapping
           consists a cycle the with existing skos:braoderTransitive relation".
validation:Pattern1NonConsistentWithSKOSRules
       rdfs:subClassOf validation:Pattern1VocabularyHijacking.
validation:Pattern1NonConsistentWithSKOSExtraRules
       rdfs:subClassOf validation:Pattern1VocabularyHijacking.
```

Listing 6 shows an extract of the classification of problematic patterns. The patterns discussed in Listing 3-5 are categorized as three rdfs:Class so as to identify the detected patterns. The non-consistent patterns stated in Listing 4 and 5 are more specific than the vocabulary hijacking pattern listed in Listing 3. If a pattern can be detected by the rules in Listing 4 or 5, it can also be detected by the rule in Listing 3 as a vocabulary hijacking. Such a relationship is reflected in Listing 6 by defining the patterns in Listing 4 and 5 as sub classes of the pattern in Listing 3. In the remainder of this section, the patterns listed in the SKOS mapping validation rules are presented.

### 3.3   PATTERN 2: Assert skos:exactMatch relation via skos:exactMatch mapping

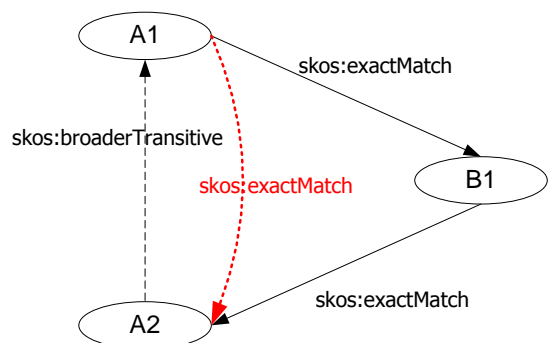

Figure 5. Validation pattern 2

Figure 5 shows the problematic pattern where a skos:exactMatch relation is asserted via the skos:exactMatch mappings. Such a relation may either result in vocabulary hijacking or create conflicts.

Listing 7. Pattern2VocabularyHijacking

```
@prefix skos: <http://www.w3.org/2004/02/skos/core#>.
@prefix validation: <http://eulersharp.sourceforge.net/2003/03swap/skos-mapping-validation-rules#>.
@prefix e: <http://eulersharp.sourceforge.net/2003/03swap/log-rules#>.
@prefix log: <http://www.w3.org/2000/10/swap/log#>.
1 {
2     ?A1 skos:exactMatch ?B1.
3     ?B1 skos:exactMatch ?A2.
4     ?A1 log:notEqualTo ?A2.
5     {    ?A1 skos:exactMatch ?B1.
6          ?B1 skos:exactMatch ?A2. } e:graphCopy ?pattern.
7 } => {
8     ?pattern a validation:Pattern2VocabularyHijacking.
9 }.
```

The triple stated on Line 4, {?A1 log:notEqualTo ?A2.}, declares ?A1 should be different from ?A2. The e:findall built-in, as what used in Line 4 of Listing 3 is not used in this rule. This is because the skos:exactMatch should not be used to indicate a semantic relation inside a concept scheme by convention.

As the skos:exactMatch is a symmetric property, after applying the prerequisite inference rules, this relation will be stated in both directions. For example, both (*A1 skos:exactMatch B1*) and (*B1 skos:exactMatch A1*) exist, though the Latter one is not explicitly displayed in Figure 5. Whenever there is a false pattern detected by the rule in Listing 7, it is therefore detected twice with both directions. As in Figure 5, both pattern (*A1 skos:exactMatch B1. B1 skos:exactMatch A2*) and pattern (*A2 skos:exactMatch B1. B1 skos:exactMatch A1.*) are detected.

Furthermore, in case there exists a semantic relation (A2 skos:broaderTransitive A1.), the inferred skos:exactMatch relation will cause a conflict with this existing relation. Such a pattern will be detected by the rule in Listing 8.

Listing 8. Pattern2NonConsistentWithSKOSExtraRules

```
{
     ?A1 skos:exactMatch ?B1.
     ?B1 skos:exactMatch ?A2.
     ?A1 log:notEqualTo ?A2.
     ?A2 skos:broaderTransitive ?A1.
     {    ?A1 skos:exactMatch ?B1.
          ?B1 skos:exactMatch ?A2.
          ?A2 skos:broaderTransitive ?A1. } e:graphCopy ?pattern.
} => {
     ?pattern a validation:Pattern2NonConsistentWithSKOSExtraRules.
}.
```

### 3.4 PATTERN 3: Assert skos:boaderTransitive relation via skos:exactMatch mapping and skos:broaderTransitive

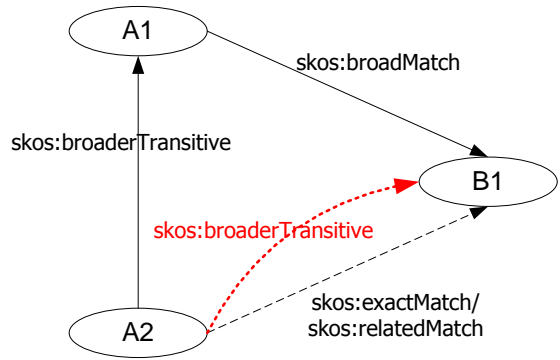

Figure 6. Pattern 3

Problematic relations are not limited to what is inferred from mapping relations. In Figure 6, the problematic relation (*A2 skos:broaderTransitive ?B1*) can be deduced with the semantic relation between A2 and A1 together with the mapping relation between A1 and B1. The inferred relation may conflict with the mapping relation between A2 and B1.

Listing 9. Pattern3NonConsistentWithSKOSRules

```
{
        ?A2 skos:broaderTransitive ?A1.
        ?A1 skos:broadMatch ?B1.
        ?B1 skos:relatedMatch ?A2.
        {       ?A2 skos:broaderTransitive ?A1.
                ?A1 skos:broadMatch ?B1.
                ?B1 skos:relatedMatch ?A2. } e:graphCopy ?pattern.
} => {
        ?pattern a validation:Pattern3NonConsistentWithSKOSRules.
}.
```

Listing 9 shows a non-consistent pattern between the inferred relation and skos:relatedMatch. Meanwhile, Listing 10 shows a non-consistent pattern between the inferred relation and skos:exactMatch.

Listing 10. Pattern3NonConsistentWithSKOSExtraRules

```
{
        ?A2 skos:broaderTransitive ?A1.
        ?A1 skos:broadMatch ?B1.
        ?B1 skos:exactMatch ?A2.
        {       ?A2 skos:broaderTransitive ?A1.
                ?A1 skos:broadMatch ?B1.
                ?B1 skos:exactMatch ?A2. } e:graphCopy ?pattern.
} => {
        ?pattern a validation:Pattern3NonConsistentWithSKOSExtraRules.
}.
```

## 3.5 PATTERN 4: Assert skos:broaderTransitive relation via skos:exactMatch mapping and skos:broaderTransitive

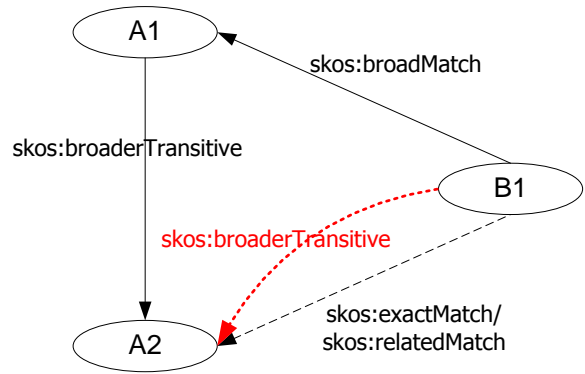

Figure 7. Pattern 4

This pattern works as the reverse direction of Pattern 3. Similarly, a skos:broaderTransitive relation can be deduced, as displayed by the red dotted line.

Listing 11. Pattern4NonConsistentWithSKOSRules

```
{
        ?A1 skos:broaderTransitive ?A2.
        ?B1 skos:broadMatch ?A1.
        ?A2 skos:relatedMatch ?B1.
        {       ?A1 skos:broaderTransitive ?A2.
                ?B1 skos:broadMatch ?A1.
                ?A2 skos:relatedMatch ?B1. } e:graphCopy ?pattern.
} => {
        ?pattern a validation:Pattern4NonConsistentWithSKOSRules.
}.
```

Listing 11 shows a non-consistent pattern between the inferred relation and skos:relatedMatch according to the SKOS-Rules. While Listing 12 shows a non-consistent pattern between the inferred relation and skos:exactMatch according to the SKOS-Extra-Rules.

Listing 12. Pattern4NonConsistentWithSKOSExtraRules

```
{
        ?A1 skos:broaderTransitive ?A2.
        ?B1 skos:broadMatch ?A1.
        ?A2 skos:exactMatch ?B1.
        {       ?A1 skos:broaderTransitive ?A2.
                ?B1 skos:broadMatch ?A1.
                ?A2 skos:exactMatch ?B1. } e:graphCopy ?pattern.
} => {
        ?pattern a validation:Pattern4NonConsistentWithSKOSExtraRules.
}.
```

## 3.6 PATTERN 5: Assert skos:broaderTransitive relation via skos:broadMatch mappings and skos:broaderTransitive

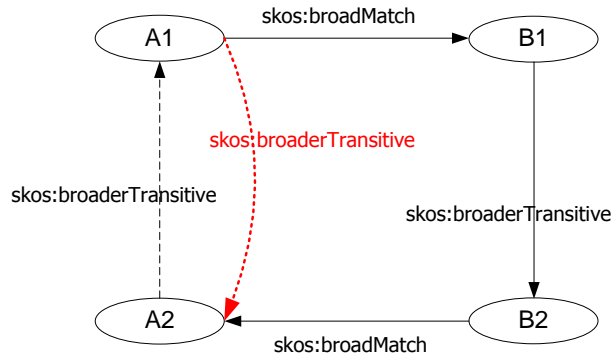

Figure 8. Pattern 5

Figure 5 shows the problematic pattern where a skos:broaderTransitive relation is asserted via the SKOS semantic and mapping relations. It may result in vocabulary hijacking or conflicts as what stated in Pattern 1.

Listing 13. Pattern5VocabularyHijacking

```
{
        ?A1 skos:broadMatch ?B1.
        ?B1 skos:broaderTransitive ?B2.
        ?B2 skos:broadMatch ?A2.
        ?A1 log:notEqualTo ?A2.
        ?SCOPE e:findall ( ?A1 { ?A1 skos:broaderTransitive ?A2. } () ).
        {       ?A1 skos:broadMatch ?B1.
                ?B1 skos:broaderTransitive ?B2.
                ?B2 skos:broadMatch ?A2. } e:graphCopy ?pattern.
} => {
        ?pattern a validation:Pattern5VocabularyHijacking.
}.
```

Listing 13 shows the inferred relation would be considered vocabulary hijacking providing it is not afore-stated.

Listing 14. Pattern5NonConsistentWithSKOSRules

```
{
        ?A1 skos:broadMatch ?B1.
        ?B1 skos:broaderTransitive ?B2.
        ?B2 skos:broadMatch ?A2.
        ?A2 skos:related ?A1.
        {       ?A1 skos:broadMatch ?B1.
                ?B1 skos:broaderTransitive ?B2.
                ?B2 skos:broadMatch ?A2.
                ?A2 skos:related ?A1. } e:graphCopy ?pattern.
} => {
        ?pattern a validation:Pattern5NonConsistentWithSKOSRules.
}.
```

Listing 14 shows a non-consistent pattern between the inferred relation and skos:related according to the SKOS-Rules. Meanwhile, Listing 15 shows a non-consistent pattern between the inferred relation and skos:broaderTransitive according to the SKOS-Extra-Rules.

Listing 15. Pattern5NonConsistentWithSKOSExtraRules

```
{
        ?A1 skos:broadMatch ?B1.
        ?B1 skos:broaderTransitive ?B2.
        ?B2 skos:broadMatch ?A2.
        ?A2 skos:broaderTransitive ?A1.
        {       ?A1 skos:broadMatch ?B1.
                ?B1 skos:broaderTransitive ?B2.
                ?B2 skos:broadMatch ?A2.
                ?A2 skos:broaderTransitive ?A1. } e:graphCopy ?pattern.
} => {
        ?pattern a validation:Pattern5NonConsistentWithSKOSExtraRules.
}.
```

## 3.7 PATTERN 6: Assert skos:broaderTransitive relation via skos:broadMatch mapping and skos:broaderTransitive

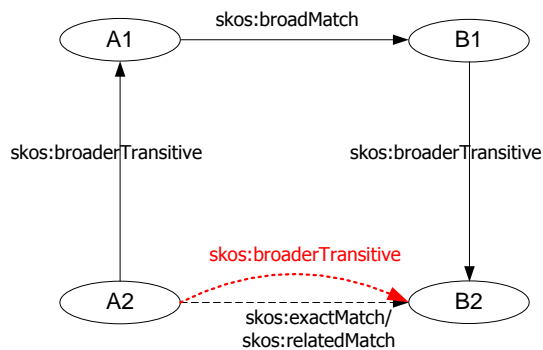

Figure 9. Pattern 6

In Pattern 6, the problematic relation (*A2 skos:broaderTransitive B2*) can be inferred, which may conflict with the mapping relations between A2 and B2.

Listing 16. Pattern6NonConsistentWithSKOSRules

```
{
        ?A2 skos:broaderTransitive ?A1.
        ?A1 skos:broadMatch ?B1.
        ?B1 skos:broaderTransitive ?B2.
        ?B2 skos:relatedMatch ?A2.
        {       ?A2 skos:broaderTransitive ?A1.
                ?A1 skos:broadMatch ?B1.
                ?B1 skos:broaderTransitive ?B2.
                ?B2 skos:relatedMatch ?A2. } e:graphCopy ?pattern.
} => {
        ?pattern a validation:Pattern6NonConsistentWithSKOSRules.
}.
```

Listing 16 shows a non-consistent pattern between the inferred relation and skos:relatedMatch according to the SKOS-Rules. Listing 17 shows a non-consistent pattern between the inferred relation and skos:exactMatch according to the SKOS-Extra-Rules.

Listing 17. Pattern6NonConsistentWithSKOSExtraRules

```
{
        ?A2 skos:broaderTransitive ?A1.
        ?A1 skos:broadMatch ?B1.
        ?B1 skos:broaderTransitive ?B2.
        ?B2 skos:exactMatch ?A2.
        {       ?A2 skos:broaderTransitive ?A1.
                ?A1 skos:broadMatch ?B1.
                ?B1 skos:broaderTransitive ?B2.
                ?B2 skos:exactMatch ?A2. } e:graphCopy ?pattern.
} => {
        ?pattern a validation:Pattern6NonConsistentWithSKOSExtraRules.
}.
```

### 3.8 PATTERN 7: Counter Intuitive

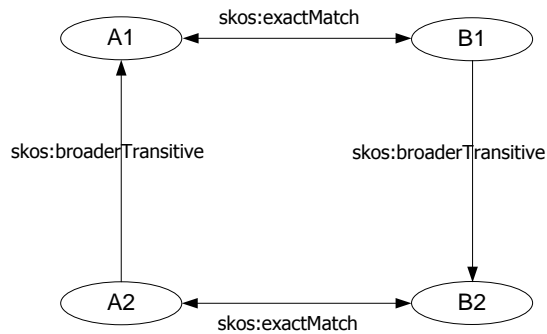

Figure 10. Pattern7

Pattern 7 neither introduces vocabulary hijacking, nor creates non-consistent patterns, according to the SKOS-Rules and SKOS-Extra-Rules. However, we consider such a mapping also problematic. The experiment in Section 4 shows an example of detecting a problematic pattern[9] as what is described in Pattern 7.

Listing 18. Pattern7CounterIntuitive

```
{
        ?A2 skos:broaderTransitive ?A1.
        ?A1 skos:exactMatch ?B1.
        ?B1 skos:broaderTransitive ?B2.
        ?B2 skos:exactMatch ?A2.
        {       ?A2 skos:broaderTransitive ?A1.
                ?A1 skos:exactMatch ?B1.
                ?B1 skos:broaderTransitive ?B2.
                ?B2 skos:exactMatch ?A2. } e:graphCopy ?pattern.
} => {
        ?pattern a validation:Pattern7CounterIntuitive.
}.
```

The patterns listed in this section described in detail what we consider problematic patterns in using SKOS mappings. The problematic patterns are categorized in three types: vocabulary hijacking, non-consistent by convention (according to SKOS-Extra-Rules), and non-consistent with the SKOS specification (according to SKOS-Rules). The detected patterns can be passed to the mapping creator. For the problematic patterns of vocabulary hijacking and non-consistent by convention, it is up to the mapping creator to decide whether to modify the mappings or to keep them as such. While for the third type, non-consistent with the SKOS specification, the mapping creator must correct the problematic mappings to avoid such non-consistency.

## 4    Experiment

The SKOS mapping validation rules discussed in this paper are published [7] and can be executed to validate SKOS mappings using the EYE reasoning engine. It is out of the scope of this paper to discuss detailed validation results of specific mappings. Nevertheless, a sample test case is created which demonstrates how a problematic pattern can be detected by the validation rules. Interested readers may find details in the following related resources. Sample mappings as well as the related concepts in both source and target concept schemes are put together in the sample data[8]. Validation results[9] are received after applying the validation rules [7] and corresponding queries[10]. Proofs[11] of those validation results are also generated. The scripts to execute the validation processes with EYE are also available[12].

## 5    Conclusions

With the development of the Semantic Web, representing terminology mappings in a semantic format is urgently required so as to better integrate data from different sources in which different terminologies are used. The SKOS vocabulary is widely used in representing terminology in a semantic format. In the SKOS vocabulary, a set of mapping properties are included to map concepts between terminologies. However, we found that due to the transitive nature of some SKOS properties, mappings expressed with SKOS mapping properties may result in problematic patterns after certain inferences. Furthermore, as most of these patterns are only visible after inferences, the mapping creators may not be aware of the existence of those problematic patterns. In addition, a valid mapping may also become invalid when the related terminologies evolve [10], especially when inference is introduced. Methods that are able to automatically detect problematic mapping patterns are therefore required in order to improve and maintain the mapping quality.

---

[8] http://eulersharp.sourceforge.net/2007/07test/skos-mapping-sample-snomed-icd10.n3
[9] http://eulersharp.sourceforge.net/2007/07test/skos_mv_answer.n3
[10] http://eulersharp.sourceforge.net/2007/07test/skos-mapping-validation-query.n3
[11] http://eulersharp.sourceforge.net/2007/07test/skos_mv_proof.n3
[12] http://eulersharp.sourceforge.net/2007/07test/skos_mv_test

In this paper, a set of SKOS mapping validation rules are developed to accomplish the aforementioned request. The rules detect problematic patterns in terminology mappings that are expressed with SKOS mapping properties. The mapping validation rules have been tested in validating mappings between medical terminologies. This paper provides a general introduction by means of a test case for using the validation rules to detect problematic patterns. Detailed analysis of the validation results of public terminology mappings in the medical domain will be presented in the future.

## Acknowledgements

This work was supported by funding from the SALUS project (http://www.salus project.eu/). Grant agreement N° 287800.